\documentclass[10pt,twocolumn,letterpaper]{article}

\usepackage{iccv}
\usepackage{times}
\usepackage{epsfig}
\usepackage{graphicx}
\usepackage{amsmath}
\usepackage{amssymb}
\usepackage{mathtools}
\usepackage{amsthm}
\usepackage{subfigure}
\usepackage[ruled,linesnumbered]{algorithm2e}

\usepackage{booktabs}
\usepackage{mathrsfs}
\usepackage{xcolor}         
\usepackage{multirow}
\usepackage{makecell}
\usepackage{wrapfig}
\usepackage{color}
\usepackage{mathabx}
\definecolor{gray}{RGB}{127,127,127}

\newcommand{\NAME}{COTI}
\newcommand{\FULLNAME}{Controllable Textual Inversion}
\newcommand{\EMPHFULLNAME}{\textbf{CO}ntrollable \textbf{T}extual \textbf{I}nversion}

\newcommand{\jz}[1]{\textcolor{yellow}{#1}}

\DeclareMathOperator*{\argmin}{arg\,min} 

\theoremstyle{plain}
\newtheorem{theorem}{Theorem}[section]

\theoremstyle{definition}
\newtheorem{definition}[theorem]{Definition}

\theoremstyle{remark}

\usepackage[breaklinks=true,bookmarks=false]{hyperref}

\iccvfinalcopy 

\def\iccvPaperID{4117} 

\ificcvfinal\fi

\begin{document}

\title{Controllable Textual Inversion for Personalized Text-to-Image Generation}

\author{Jianan Yang, Haobo Wang, Yanming Zhang, Ruixuan Xiao, Sai Wu, Gang Chen, Junbo Zhao\\
$\phantom{2}^{1}$College of Computer Science and Technology, Zhejiang University\\
{\tt\small \{jianan0115,wanghaobo, yanmingzhang, xiaoruixuan,wusai,cg,j.zhao\}@zju.edu}
}

\maketitle
\ificcvfinal\thispagestyle{empty}\fi

\begin{abstract}
The recent large-scale generative modeling has attained unprecedented performance especially in producing high-fidelity images driven by text prompts.
Text inversion (TI), alongside the text-to-image model backbones, is proposed as an effective technique in personalizing the generation when the prompts contain user-defined, unseen or long-tail concept tokens.
Despite that, we find and show that the deployment of TI remains full of ``dark-magics'' --- to name a few, the harsh requirement of additional datasets, arduous human efforts in the loop and lack of robustness.
In this work, we propose a much-enhanced version of TI, dubbed \EMPHFULLNAME{} (\NAME{}), in resolving all the aforementioned problems and in turn delivering a robust, data-efficient and easy-to-use framework.
The core to \NAME{} is a theoretically-guided loss objective instantiated with a comprehensive and novel weighted scoring mechanism, encapsulated by an active-learning paradigm.
The extensive results show that \NAME{} significantly outperforms the prior TI-related approaches with a 26.05 decrease in the FID score and a 23.00\% boost in the R-precision.
\end{abstract}

\begin{figure*}[t]
\centering
\includegraphics[width=1.\textwidth]{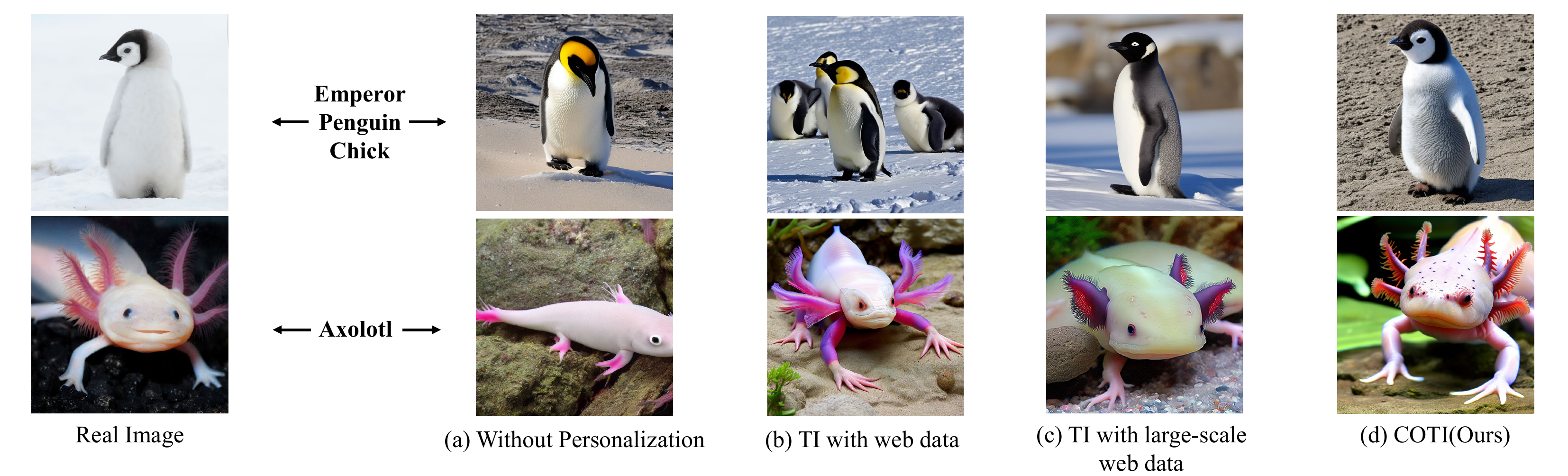}
\caption{\small \textbf{Comparison of generated images on different versions of personalized text-to-image generation.} This figure shows a comparison of (a) images generated without textual inversion embedding, (b) images generated with TI trained on 100 randomly selected web data, (c) images generated with TI trained on 1000 web data, (d) images generated with \NAME{} with 100 automatically-selected data. The experiments are conducted on the publicly available Stable Diffusion 2.0 with the concepts “emperor penguin chicks” and “axolotl”.
}
\label{fig:intro_example_comparison}
\end{figure*}

\section{Introduction}

The generative modeling for text-to-image has attained unprecedented performance most recently~\cite{nichol2021glide,rombach2022high,ramesh2022hierarchical,saharia2022photorealistic}.
Notably, by training over billions of text-image data pairs~\cite{schuhmann2021laion,schuhmann2022laion}, the family of diffusion models has allowed high-fidelity image synthesis directed by the \emph{prompt} provided in production.
Despite their massive successes, these models still fail to generate images with decent quality and matching semantics, when encountering prompts that contain unseen or long-tailed concept tokens~\cite{skantze2022collie,ding2022don}.
Simply put, the reason causing this failure is due to that the training set is \textbf{not} unbounded with limited variations.
On the side of production, this drawback may severely obstruct the \emph{personalization} of this technique towards scaled deployment.

Text inversion (TI)~\cite{gal2022image} has been proposed recently to address such personalization problems. 
Briefly, TI facilitates an insertion mechanism to the embedding space while having the rest of the model untouched and frozen. 
It is generally trained by gradient descent merely on the inserted embedding vectors, towards reducing the loss calculated on some newly-curated paired data containing the targeted concept tokens.
As a result, the inserted embeddings are dedicated to representing the targeted concept tokens that the model originally does not perceive adequately.
Unlike the other common paradigms like re-training or fine-tuning~\cite{ding2022don,kumar2022fine,cohen2022my}, TI avoids the catastrophic forgetting problem, nor does it incur prohibited computation.

Despite the promise, however, the practical shipping and conduction of the TI framework are not foolproof.
As we show in Fig.~\ref{fig:intro_example_comparison}~(b), training the TI embedding by default using large-scale web-crawled data does not yield the expected image quality.
Unfortunately, most --- if not all --- of the related prior work rely on human annotation or data selection~\cite{ding2022don,li2022overcoming,zhou2022learning,skantze2022collie}.
However, the involvement of human annotation may crucially limit the versatility of the TI framework and somewhat contradicts its design principle of being a lightweight and fast paradigm~\cite{li2020personality,he2022rethinking}.

In this article, we comprehensively study the aforementioned problem.
The ultimate goal of this work is to deliver an \textbf{active and controllable data selection framework} for enhancing the vanilla TI framework. 
The major course of this work is to substitute the human labeling/querying procedure with a holistic and automated mechanism, which has not been discussed or addressed in prior work.
In this regard, we propose \EMPHFULLNAME{}, shorted as \NAME{}.
The profound motivation of this approach is to comprehensively model the interaction between the online paired data with the embedding space of the Stable diffusion model~\cite{rombach2022high}.
In particular, we devise a progressive and active-learned paradigm catered to TI that dynamically expands the training data pool, by contrast to the prior work using a fixed data pool and sampling/training schedule.
Indeed, the traditional active learning scheme aims to reduce the quantity of data required to yield a decent model. 
However, in this scenario, we found that simply enlarging the training pool by gathering more web-crawled data does not provide a noticeable gain, shown in Fig.\ref{fig:intro_example_comparison}~(c). 
As such, we argue that the actively drawn sample pool in \NAME{} should guarantee both their quality and quantity.

Diving into the construction of \NAME{}, we further devise a novel \emph{scoring system} that evaluates each queried data pair.
In spite of the few literature work on image quality evaluation~\cite{ren2021survey,herde2021survey,he2022rethinking,chen2022deep}, our scoring system differs in two major aspects: (i)-this scoring system combines all major concerns for evaluating image quality and (ii)-the scoring system dynamically adapts itself alongside the update of the embedding.
In other words, directly applying the prior scoring system to TI does not suffice, because the evaluated scores may not necessarily contribute or align with the altered embedding.
We provide our theoretical guidance for scoring with MMSE in Sec.~\ref{sec:method} and discuss this phenomenon in detail in Sec.~\ref{sec:exp}.

To summarize, we provide a comprehensive exhibition of the image generated with or without TI, compared with \NAME{}, in Fig.~\ref{fig:intro_example_comparison}.
We may conclude that our approach \NAME{} powered by automatically selected high-quality data yields the best quality, fidelity and semantic alignment, without any interference from manual efforts.
The profound contribution of this article can be summarized as follows: we develop a holistic and much-enhanced text inversion framework that achieves significant performance gain with \textbf{26.05} on FID score, \textbf{23.00\%}  on R-precision. 

\begin{figure*}[t]
\centering
\subfigure[Outline of the Textual Inversion Process.]{\includegraphics[width=0.47\textwidth]{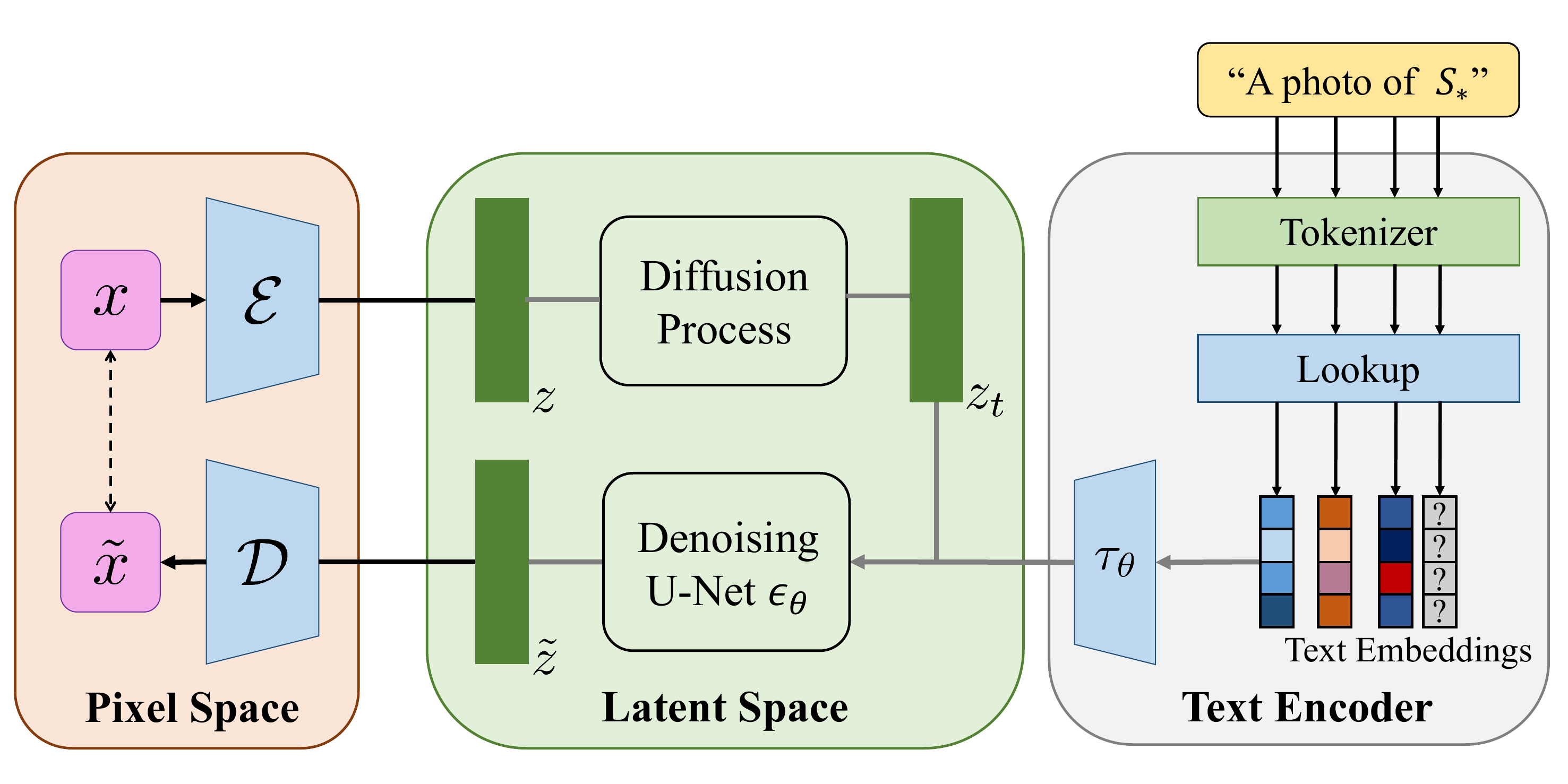}\label{fig:ti}}
\subfigure[Outline of the \FULLNAME{} Process.]{\includegraphics[width=0.47\textwidth]{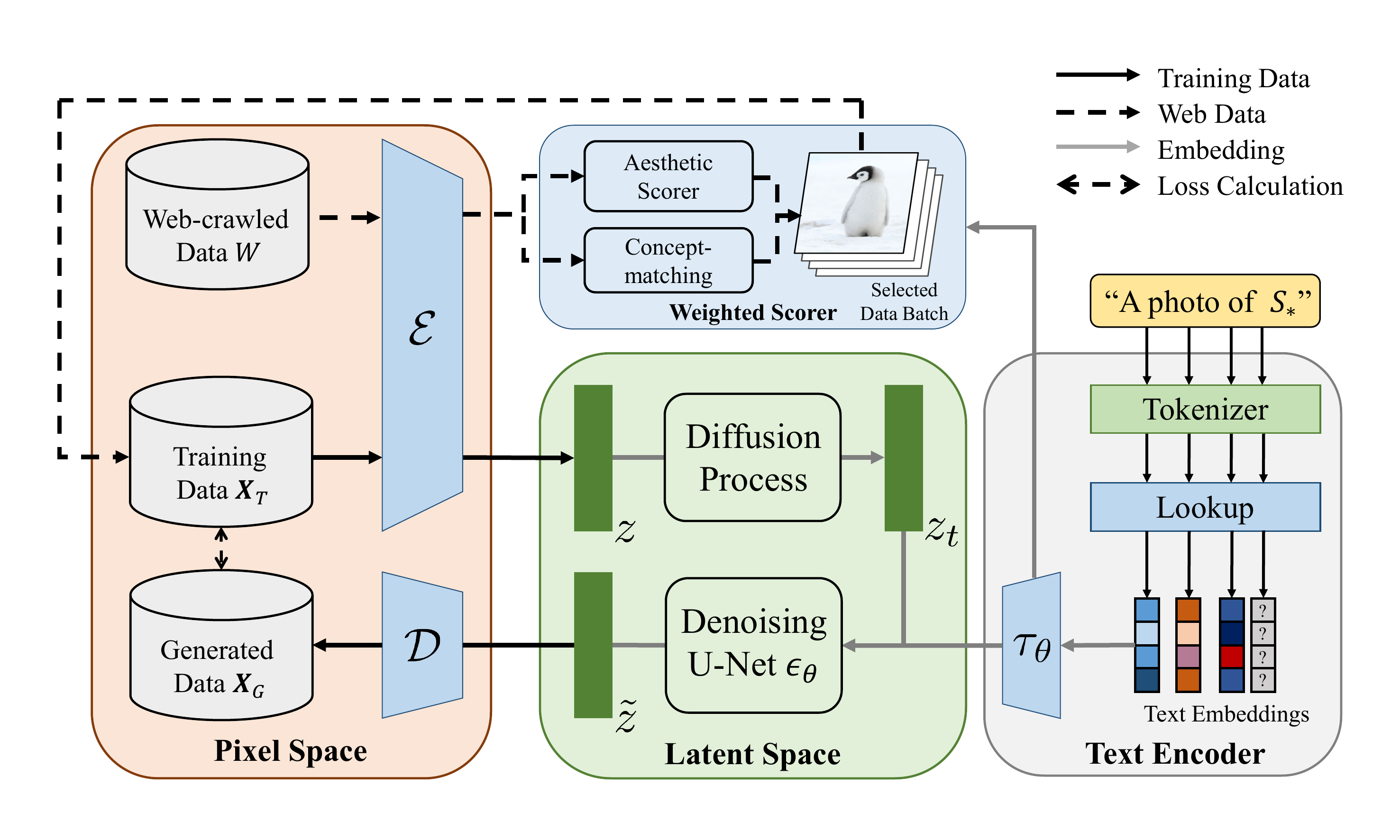}\label{fig:dati}}
\caption{
\textbf{Left: } Vanilla TI learns from a high-quality but small image set $\mathbf{X}_T$. 
\textbf{Right: } \NAME{} alternatively perform data selection and scheduled embedding training. In each training cycle, \NAME{} calculates a weighted score based on the current state of text embedding and then selects new samples to expand $\mathbf{X}_T$. 
Afterward, the text embedding is trained via a dynamically scheduled procedure. 
The above two steps alternatively proceed until convergence. 
We show the approach to apply \NAME{} to a diffusion-based text-to-image model. 
} 


\label{fig:intro_ti_dati_comparison}
\end{figure*}

\section{Background}

Textual inversion~(TI)~\cite{gal2022image} is a learning paradigm especially designed for introducing a new concept into large-scale text-to-image models, in which the concept is originally not involved in the model. 
It learns an extra text embedding corresponding to the newly-given concept, which can be further used to guide the generative model to synthesize images of this concept.
Basically, textual inversion is built upon Latent Diffusion Models~(LDMs)~\cite{rombach2022high}, which comprise two components: an auto-encoder consisting of an encoder $\mathcal{E}$ and a decoder $\mathcal{D}$ and a diffusion process operating the latent space. 
Furthermore, the diffusion process can be conditioned on the output of text embedding models, enabling the auto-encoder to integrate the information derived from texts. Let $c_{\theta}$ the text encoder that maps the text $y$ into this conditioning vector, the LDM loss is:
\begin{equation*}
    L_{LDM}\coloneq\mathbb{E}_{z\sim\mathcal{E}(x),v,\epsilon\sim\mathcal{N}(0,1),t}\left[\Vert 
\epsilon - \epsilon_\theta\left(z_t,t,c_\theta (y)\right) \Vert_2^2\right],
\label{eq:ldm_loss}
\end{equation*}
where $t$ denotes the time step, $z_t$ denotes the latent noised at time step $t$, $\epsilon,\epsilon_\theta$ represents the noised samples and the denoising network, respectively.

Based on LDM, textual inversion focus on giving an appropriate conditioning vector generated by text embedding models.
Typical text encoding models, like BERT~\cite{devlin2018bert}, convert a word into a token and associate the token with a unique embedding vector that can be retrieved through an index-based lookup, in which the embedding vector is learned as a part of the text encoder $c_\theta$. Specifically, textual inversion focuses on optimizing a specific text embedding that corresponds to the newly given concept, with all the other components and embeddings frozen. The optimization goal of TI can be formulated as follows:
\begin{equation}
    v^*=\argmin_v \mathbb{E}_{z\sim\mathcal{E}(x),v,\epsilon\sim\mathcal{N}(0,1),t}\left[\Vert 
\epsilon - \epsilon_\theta\left(z_t,t,\tau_\theta(v)\right) \Vert_2^2\right],
\label{eq:textual_inversion}
\end{equation}
in which we use $v$ to denote the unique embedding vectors corresponding to the newly-given concept when $\tau_{\theta}$ denotes the transformer that converts text embedding into the latent space of LDM. An outline of TI is also given in Fig.~\ref{fig:ti}.
In our work, we denote the textual inversion loss $L_{LDM}$ by $L_{LDM}(x,v)$ and simply denote the generative model with $g(x;v)$ in the following contexts.

\section{Method}\label{sec:method}

In this section, we elaborate \NAME{} framework.
To begin with, we outline the overall framework in Sec.~\ref{sec:main_process}.
In what follows, we detailed the key components --- the actively-learned data selection process corporated with the dual-scoring system, in Sec.~\ref{sec:dual_scoring_system}.
At last, we provide some important technical desiderata towards implementation for \NAME{}.
An overview of \NAME{} is illustrated in Fig.~\ref{fig:dati}.


\subsection{\FULLNAME{}}\label{sec:main_process}
We write down the objective in the form below:
\begin{equation}
    v^*, \mathbf{X}_T^* = \argmin\nolimits_{v, \mathbf{X}_T}\mathbb{E}_{x\sim \mathbf{X}_T}L_{LDM}(x,v),
    \label{eq:ultimate_goal}
\end{equation}
where the vector $v$ denotes the text embedding we want to learn, while the random vector $\mathbf{X}_T$ represents the training set.
Noted, compared with Eq.~\eqref{eq:textual_inversion} for the vanilla form of text inversion, we proactively integrate the dataset distribution $\mathbf{X}_T$ into the objective.
To optimize it, we decompose this objective into the following dual conjugate forms:
\begin{align}
 \mathbf{X}_T^{(t)} &= \argmin\nolimits_{\mathbf{X}_T\in \mathcal{D}(W)}\mathbb{E}_{x\sim \mathbf{X}_T}L_{LDM}(x,v), \label{eq:iterative_dati_data} \\
 v^{(t)} &= \argmin\nolimits_{v}\mathbb{E}_{x\sim \mathbf{X}_T^{(t-1)}}L_{LDM}(x,v),
\label{eq:iterative_dati_embedding}
\end{align}
where $\mathcal{D}(W)$ is the web data distribution, $t$ denotes the cycle step. $\mathbf{X}_T^{(0)}$ is a manually selected high-quality sample set, and we iteratively expand the training set $\mathbf{X}_T$ beginning from $\mathbf{X}_T^{(0)}$.
After the transmission, the learning procedure of \NAME{} is relatively clear: the sample pool is progressively accumulated --- by Eq.~\eqref{eq:iterative_dati_data} --- and the optimization of the new embedding vector $v$ is straightforward.

The only remaining problem concerns the enrichment of the paired training set $\mathbf{X}_T$.
Essentially, this pool of paired data is constructed and enriched from the web-crawled dataset $W$ under certain criteria. As we mentioned, we would like to collect high-quality samples that efficiently reduce the loss term in Eq.\eqref{eq:iterative_dati_data}. 
However, directly optimizing $\mathbf{X}_T$ is not feasible, because the search space of such combinatorial distribution expanded from $\mathcal{D}(W)$ is prohibitively intractable~(up to $2^{|W|}$). In what tags along, the unit computational cost of optimizing $L_{LDM}$ via the large-scale text-to-image on each possible set $\mathbf{X}_T$ is not negligible.
Therefore, to appropriately optimize Eq.~\eqref{eq:iterative_dati_data}, we ought to propose a novel mechanism that rapidly estimates the impact of each sample on the LDM loss term with manageable computational cost.
To achieve this, we first draw inspiration from the MMSE theories that estimate a specific random vector from other observation inputs: 
\begin{definition} The \textit{minimum mean square error (MMSE)} of estimating an input random vector $\widehat{\mathbf{X}}\in\mathbb{R}^n$ from an observation/output $\mathbf{X}\in\mathbb{R}^k$ is defined as
\begin{equation}
    \mathrm{MMSE}( \widehat{\mathbf{X}}|\mathbf{X} ) = \inf\nolimits_{f\in\mathcal{M}(\mathbb{R}^n)} \mathbb{E} \left[ \left\Vert \widehat{\mathbf{X}} - f(\mathbf{X}) \right\Vert^2 \right],
    \label{eq:mmse_def}
\end{equation}
in which $\mathcal{M}(\mathbb{R}^n)$ denotes the space consisting of all measurable functions on $\mathbb{R}^n$.
\label{definition:mmse}
\end{definition}

The definition above shows how we estimate an unknown distribution from observations by minimizing an MMSE term. Notice that we are trying to produce the best generation results which are initially unknown, our personalization task can be also regarded as estimating the optimal distribution with carefully selected data. By denoting the generative results and training distribution with random variables $\widehat{\mathbf{X}}_G$ and $\mathbf{X}_T$, we conclude that the LDM loss $L_{LDM}$ is consistent with this MMSE term:
\begin{theorem}
    Let $L_{LDM}$ be the LDM loss following Eq.~\ref{eq:ldm_loss}, and the image space lies within $\mathbb{R}^n$. 
    Then there always exists an ideal random vector $\widehat{\mathbf{X}}_G\in\mathbb{R}^n$ and a condition vector $v^*$ within the text embedding space for LDM, such that
    \begin{equation*}
    \resizebox{1.0\linewidth}{!}{$\begin{aligned}
        \argmin_{\mathbf{X}_T\in\mathbb{R}^n}\mathrm{MMSE} (\widehat{\mathbf{X}}_G|\mathbf{X}_T) = \argmin_{\mathbf{X}_T\in\mathbb{R}^n}\mathbb{E}_{x\sim \mathbf{X}_T}\left[L_{LDM}\left(x,v^*\right)\right].\end{aligned}$}
    \end{equation*}
    \label{theorem:ldm}
\end{theorem}
\noindent Computationally, the changes given by newly-added samples on the MMSE term can be estimated through a scoring system, which we describe in the following section.

\subsection{Weighted Scoring System}\label{sec:dual_scoring_system}

Based on the prior deduction, we hereby introduce the scoring system built in \NAME{}.
As we mentioned, the scores are proximal form deduced by MMSE losses to assess the samples' contribution towards reducing the original LDM loss term.
To quantify the extent of such contribution, we further draw inspiration from the following theorem:
\begin{theorem} (Pythagorean Theorem for MMSE~\cite{dytso2018structure}.) Following Theorem~\ref{theorem:ldm}, by setting $f$ to the generative model $g(v)$, the minimum mean square error (MMSE) in Eq.~\eqref{eq:mmse_def} can be decomposed into two terms:
\begin{equation}
    \begin{aligned}
    \mathbb{E}\left[\left\Vert\widehat{\mathbf{X}}_G - g(v^*)\right\Vert\right] &= \mathbb{E}\left[\left\Vert\widehat{\mathbf{X}}_G-\mathbb{E}\left[\mathbf{X}_G|\mathbf{X}_T\right]\right\Vert\right] \\
    &+\mathbb{E}\left[\left\Vert g(v^*) - \mathbb{E}\left[ \mathbf{X}_G|\mathbf{X}_T\right] \right\Vert\right],
    \end{aligned}
    \label{eq:pythagorean}
\end{equation}
\label{theorem:mmse_pythagorean}
\end{theorem}
\noindent Here $\widehat{\mathbf{X}}_G$ denotes an ideal distribution for the generated results, and $g(v^*)$ represents the text-to-image generative results driven by the prompt containing the new concept. 
On the right-hand side, $\mathbb{E}\left[\mathbf{X}_G | \mathbf{X}_T\right]$ points to the produced images conditioned on the training set or its representations, i.e. the output of the image autoencoder. 
By carefully revisiting the two terms, we may further decompose the MMSE into two components, as follows.

\paragraph{The first term $\rightarrow$ Aesthetics.}
Intuitively, the first term aims to estimate how far the generative distribution $\mathbf{X}_G|\mathbf{X}_T$ drifts away from the ideal distribution of $\widehat{\mathbf{X}}_G$. 
Nevertheless, since this ideal distribution is ubiquitously unobservable or unknown and thus unoptimizable, we connect it to how humans may perceive the ideally-generated images in reality.
We provide a practical instantiated form to quantify this -- aesthetic score.
We follow the most straightforward solution by fine-tuning an off-the-shelf aesthetic perception model~\cite{deng2017image,jang2021analysis,yang2022personalized}. 
Formally, we denote the prediction by:
\begin{equation}
    p=S_{aes}(x,\theta),\label{eq:aesthetic_score}
\end{equation}
where $S_{aes}$ indicates the aesthetic scoring model parametrized by $\theta$, $p$ and $x$ denote the predicted score and image respectively.
In \NAME{}, this aesthetic score --- measuring how human aesthetically perceives the generated images --- effectively serves as a surrogate of assessing the gap towards the ideal generated distribution.
Notably, by fine-tuning $\theta$ over specific paired data, we can easily extend it to other unseen concepts.




\paragraph{The second term $\rightarrow$ Concept-matching.}
Now let us take a look at the second term.
In hindsight, this term diminishes when the unseen concept token gets perfectly reflected in the generated results.
Minimizing it amounts to reducing the semantic gap between the functional reflection of the provided unseen target. 
However, the ideal text embedding $v^*$ is unknown initially, making it impossible to induce a scoring function. Fortunately, the original $\mathbf{X}_T$ is manually selected and can reflect human preferences for ideal concept embeddings. Hence, we resort to estimating the following surrogate form:
\begin{equation}
    S_{con}(x)=\min\nolimits_{x'\in \mathbf{X}_T}\text{cosine}(\mathcal{E}(x), \mathcal{E}(x')),
    \label{eq:concept_matching}
\end{equation}
where $\mathcal{E}$ is the image encoder within the text-to-image model.
Essentially this form quantifies the gap between the embedding-induced images against their good counterparts drawn from the web.
Similar techniques are proposed from separate research domains such as image-retrieval~\cite{chen2022deep}.
To sum up, we use the $S_{con}(x)$ to score the extent of concept-matching of the image $x$.

\begin{algorithm}[t]
	\caption{The paradigm of \NAME{}.}
	\label{alg:cycle_dati}
	\KwIn{
                Web data pool $W$,
                training pool $\mathbf{X}_T$, 
                query batch size $B$,
                learning rate group $R=\{r_1,\dots,r_m\}$,
                cycles $N$.}
	\KwOut{New concept embedding $v$.}  
	\BlankLine
 
        


        \For{$cycle\gets 1$ \KwTo $N$}{
            Select $B$ samples from $W$ with highest $S_{int}$ (Eq.~\eqref{eq:score_integration}) and add $B$ to $\mathbf{X}_T$;
            
            \For{$r\gets r_1,\dots,r_m$ within $R$ in decreasing order}{
                Train the embedding $v$ on $\mathbf{X}_T$ with learning rate $r$;
                
                Repeat training until $\gamma(v)$ declines;
            }
        }

\end{algorithm}

\subsection{Practical Implementation}\label{sec:dynamic_scheduling}
In this section, we provide the specific instantiation of \NAME{} with some implementation choices.
Specifically, the overall training scheme of \NAME{} lies in an active learning framework, where the training of the embedding and the accumulation of the dataset alternatively proceed cycle-by-cycle.
The main reason for using the active learning paradigm is its preferred sample efficiency, with better controllable data bias management~\cite{farquhar2021statistical,sapkota2022balancing}.
Algorithm~\ref{alg:cycle_dati} illustrates the overall pipeline of our method.



\paragraph{Trading-off the two scores.}

Despite that we manage to decompose the objective into two interconnected scores, how to combine the two scores for sample selection and training remains a realistic implementation problem.
If we look closely, the two scores are irrelevant to the concept embeddings,
because $S_{aes}$ concerns the current state of the model and input images (Eq.~\eqref{eq:aesthetic_score}) while $S_{con}$ emphasizes a decent reference dataset tagged by the targeted concept (Eq.~\eqref{eq:concept_matching}).
Empirically we found that simply adding them, or balancing them with a fixed coefficient, does not yield the expected results.
Therefore, thanks to the active learning paradigm, \NAME{} allows a more flexible and versatile weighting strategy based on the current state of concept embeddings.

We write down the balancing coefficients as follows:
\begin{align}
    \gamma_{aes}(v) &= \frac{1}{|\mathbf{X}_{TI}|}\sum\nolimits_{x_g\in\mathbf{X}_{TI}}S_{aes}(x_g,\theta), \label{eq:vec_aes}\\ 
    \gamma_{con}(v) &= \frac{1}{|\mathbf{X}_{TI}|}\sum\nolimits_{x_g\in\mathbf{X}_{TI}}S_{con}(x_g),\label{eq:vec_con}
\end{align}
where $\mathbf{X}_{TI}$ denotes the generated image set guided by the text embedding obtained at the last active cycle. And, in its turn an integrated score:
\begin{equation}
    \resizebox{1.0\linewidth}{!}{$\begin{aligned}
        S_{int}(x) = \left(1-\frac{\gamma_{aes}(v)}{10}\right)S_{aes}(x) + \left(1-\frac{\gamma_{con}(v)}{10}\right)S_{con}(x).\end{aligned}$}\label{eq:score_integration}
\end{equation}
This formulation offers some meritable advantages. On one hand, the balancing terms are not pre-fixed or manually tuned, but dynamically dependent on the scores marginalized over the current generations. Further, the weights are monotonically disproportionate to the scores respectively; for instance, when the score of either side gets larger, the corresponding balancing coefficient in turn decreases. As a result, this mechanism encourages an alternation of sample selection towards both scoring metrics, which effectively guarantees the sample set diversity.

\paragraph{Training schedule be dynamic.}
The learning rate schedule has been found crucial for successful text inversion, and the vanilla TI method typically relies on manually designed decay rules. In \NAME{}, since the training data dynamically changes, we empirically find the text embeddings may overfit and be biased toward on dominant score without careful scheduling, which we empirically show in Fig.~\ref{fig:schedule_score_compare}. 
To remedy this problem, we establish a new \emph{indicator} that measures the changes in both scores:
\begin{equation}
    \phi(v)=10\sin\left(\frac{\pi}{20}\gamma_{aes}\left(v\right)\right)\sin\left(\frac{\pi}{20}\gamma_{con}\left(v\right)\right), \label{eq:embedding_final_score}
\end{equation}
that $\phi(v)$ depends on a trigonometric functional form of both scores.

In training \NAME{}, we use this indicator for learning-rate decay and early stopping.
Noticeably, the indicator $\phi(v)$ peaks when the two scores get close, and it bottoms when the scoring exhibits a stronger signal of being biased (to either side).
Thus, we may use it as a signaling proxy throughout the training process.
In our implementations, we calculate $\phi(v)$ every 100 epochs. When this indicator gets lower than the last evaluation, we lower the learning rate and freeze the sample pool for 100 additional epochs, in order to let the model fit more thoroughly.
By contrast, when we find the indicator hits a high plateau for several rounds, \NAME{} facilitates a new cycle of sample selection and then proceeds with training upon the extended dataset. 

\begin{figure*}[t]
    \centering
    \includegraphics[width=0.85\textwidth]{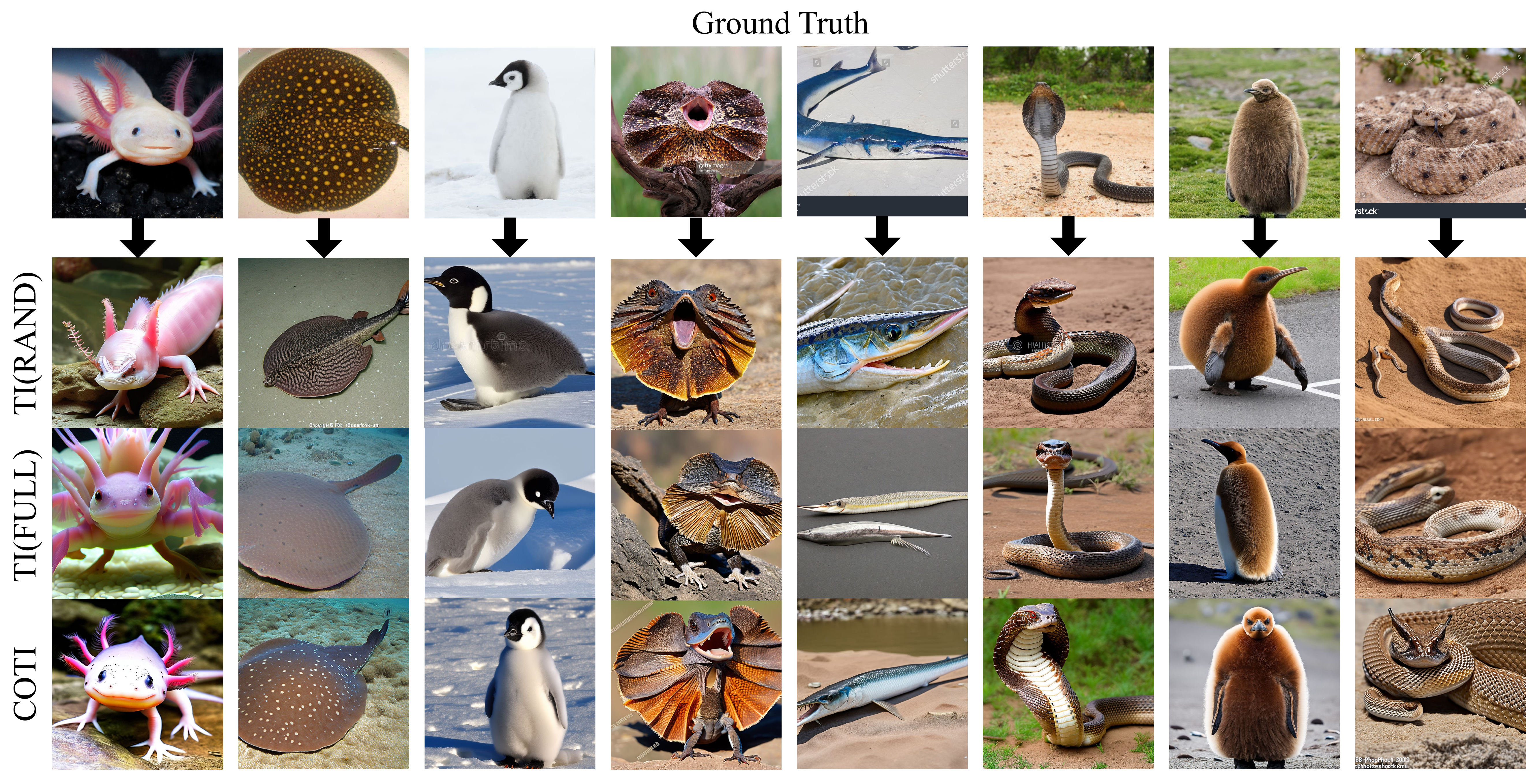}
    \caption{\textbf{A comparison of different frameworks.} Specifically, we compare three lines of works: (1) TI (RAND), in which the embedding is trained with TI and 200 randomly selected samples; (2) TI trained with all data~(1000 samples) within the dataset; (3) \NAME{} trained with carefully selected data with a budget of 200. Embeddings obtained from randomly sampled data fail to produce the features for the newly-given concepts, while those trained on full data contain necessary details but inevitably contain disruptive details.}
    \label{fig:exp_example_comparison}
\end{figure*}

\begin{figure*}[t]
    \centering
    \includegraphics[width=0.85\textwidth]{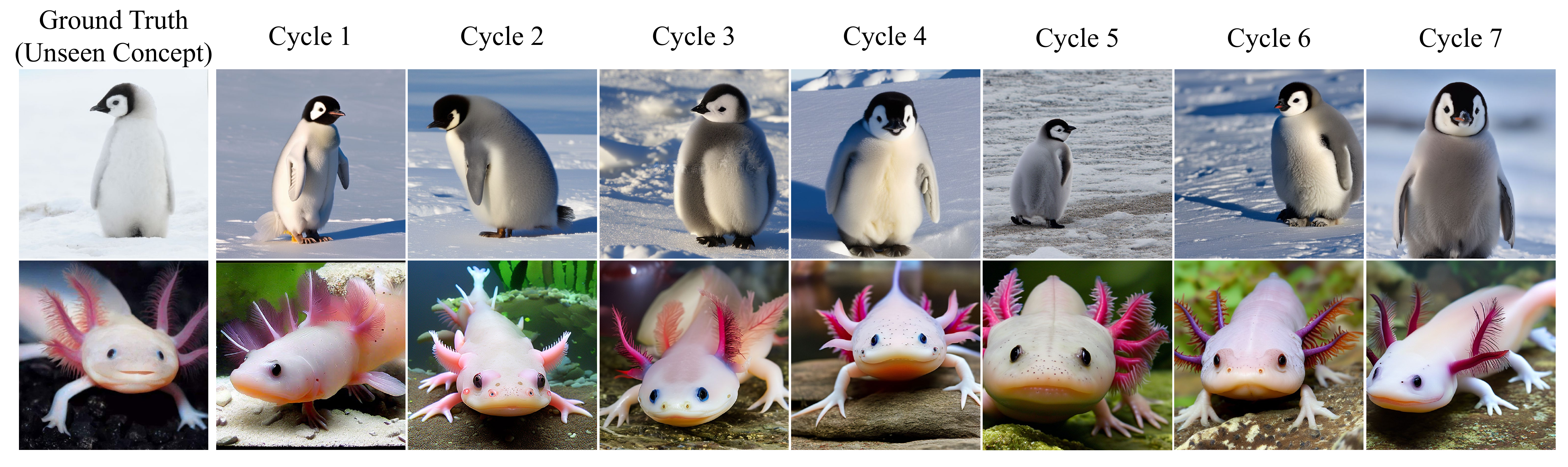}
    \caption{\textbf{Generative results as cycle proceeds}. Samples are generated with \NAME{} on cycles from 1 to 7. In each cycle, we select and add 10 high-quality samples. Generative samples start to converge and contain the right details within the original concept after cycle 4 or 5. We can also see that those generative results contain diverse contents within the background based on the few images given.}
    \label{fig:exp_example_comparison_cycle}
    \vspace{-1em}
\end{figure*}

\section{Experiments}\label{sec:exp}
In this section, we present the main experimental results both qualitatively and quantitatively. More experimental results can be found in Appendix. \textit{The source code is attached in the Supplementary. }


\paragraph{Datasets. } We trained our embeddings on datasets with 8 concepts that can hardly be generated through prompt engineering on the text-to-image model. The dataset is collected from publicly available datasets including ImageNet, iNaturalist 2018~\cite{van2018inaturalist}, IP102~\cite{wu2019ip102}, together with 1000 web-crawled data collected according to the keywords. We use the text-to-image the commonly adopted Stable Diffusion 2.0 pre-trained on LAION-5B~\cite{schuhmann2022laion} following Rombach's work~\cite{rombach2022high}. See the Appendix for detailed implementation and evaluations.

\subsection{Qualitative Results of \NAME{}}\label{exp:qualitative_results}

\paragraph{Comparison with TI.} We qualitatively compare \NAME{} with TI on their generated images. Specifically, there are two versions of TI: (1) TI (RAND), in which TI is trained with randomly sampled 200 data~(the same as the budget for \NAME{}), (2) TI (FULL), in which TI with all the 1,000 web data given. The experiments are conducted on 8 different concepts, as shown in Fig.~\ref{fig:exp_example_comparison}.
With TI (RAND), we observe that the generative results only partially learned some photographic attributes like color and texture, but fail to make the objects have the correct appearance and shape.
TI (FULL) somehow produces the corrected shape of the concept, but still fails on capturing the necessary details for describing the object, which also makes generated images unmatched by the concept.
In comparison, our proposed \NAME{} consistently match all the photographic attributes across all the given concepts while also guaranteeing the image aesthetics, making them extremely high-fidelity just like the real-world photos.

\paragraph{Generative results across cycles.} To further look at how \NAME{} learns the photographic attributes that precisely match the concept, we show samples generated from different cycles in Fig.~\ref{fig:exp_example_comparison_cycle}.
We can see that some attributes like color, and texture are already learned in cycle 1, but the shape of the object does not match the ground-truth ones. 
From cycle 1 to 3, \NAME{} shows a clear shape modification, making the objects more like the real ones.
From cycles 3 to 5, an iterative refinement on more photographic details like light, contrast, and other minor modifications~(like the peak for penguins and antenna for axolotls) is shown in generative results, making them hard to distinguish from the ground-truth ones even with a careful look.
At cycle 5 and later cycles, the image quality stabilizes and we can hardly see enhancement apart from image diversity.
To conclude, we can see an explicit attribute matching process from easy ones to the finer ones within \NAME{}, showing the importance and effectiveness of iterative training and refinement.

\begin{table*}[t]
    \centering
    \caption{\textbf{Experimental results of different strategies on 8 concepts}. We compare our dual selective active training strategy with random selection and with a full dataset. The best results are highlighted in bold-fold.}
    \vspace{0.5em}
    \resizebox{0.8\textwidth}{!}{
    \begin{tabular}{l|ccc|ccc}
        \toprule
        \multirow{2}{*}{Concepts} & \multicolumn{3}{c|}{Performance(FID$\downarrow$)} & \multicolumn{3}{c}{Performance(R-precision(\%)$\uparrow$)}  \\ \cmidrule{2-7}
                                & TI (RAND) & TI (FULL) & $\phantom{23}$\textbf{\NAME{}}$\phantom{23}$ & TI (RAND)         & TI (FULL)         & $\phantom{23}$\textbf{\NAME{}}$\phantom{23}$ \\ \midrule
         Axolotl                & 64.60$\phantom{2}$ & 55.47$\phantom{2}$ & \textbf{37.75$\phantom{2}$} & 69.56    & 72.57    & \textbf{92.36}   \\
         Frilled lizard         & 61.43$\phantom{2}$ & 44.72$\phantom{2}$ & \textbf{32.77$\phantom{2}$} & 62.33    & 68.39    & \textbf{91.39}    \\
         Crampfish              & 97.34$\phantom{2}$ & 86.39$\phantom{2}$ & \textbf{72.28$\phantom{2}$} & 72.35    & 76.56    & \textbf{94.23}    \\
         Garfish                & 136.95 & 117.82 & \textbf{104.09} & 75.62    & 75.23    & \textbf{90.26}    \\
         Emperor penguin (chick) & 144.84 & 120.37 & \textbf{94.32$\phantom{2}$} & 68.94    & 70.03    & \textbf{88.06}    \\
         King penguin (chick)    & 128.97 & 110.32 & \textbf{90.45$\phantom{2}$} & 65.65    & 64.14    & \textbf{86.23}   \\
         Indian cobra           & 120.78 & 108.23 & \textbf{92.67$\phantom{2}$}  & 70.26    & 72.35    & \textbf{87.56}   \\
         Sidewinder             & 137.56 & 121.89 & \textbf{108.33} & 68.39    & 69.12    & \textbf{85.24}    \\ \bottomrule
    \end{tabular}}
    \label{tab:main_score}
\end{table*}

\subsection{Quantitative Results of \NAME{}}\label{exp:evaluation}

\paragraph{\NAME{} obtains better generative performance.}
We evaluate the quality of our generated images with the widely used FID~\cite{heusel2017gans} and R-precision~\cite{xu2018attngan}, which quantify the embedding quality in two aspects. Specifically, the FID score measures the similarity between generated images with the real ones, indicating better generative results with lower values. Meanwhile, R-precision evaluates how generated images match a given text, which is expected to be as high as possible.

Table \ref{tab:main_score} reports FID scores and R-precision from each strategy with embeddings trained on 8 different concepts and evaluated through the generative results guided by embeddings. Both image-matching and text-matching scores exhibit better results compared to the baselines suggesting that our proposed \NAME{} significantly outperforms other frameworks, which is consistent with the qualitative results shown in Fig.~\ref{fig:exp_example_comparison}. In addition, we can see that the FID score has a large value and varies across different concepts, as our primary goal is to train a text embedding to guide the generation of this concept that can be integrated into different scenarios, rather than reconstructing images. 


\begin{table}[t]
    \centering
    \caption{\textbf{Comparing \NAME{} with other single-score based strategies.} We apply a dynamic training schedule to all the strategies here for a fair comparison. The results are produced with 6 single-score strategies adopted from evaluation metrics for image generation tasks, and aesthetic-based or retrieval-based scores.} 
    \vspace{0.5em}
    \resizebox{0.7\linewidth}{!}{
    \begin{tabular}{c|cc}
        \toprule
        Method       & FID              & R-precision (\%) \\ \midrule
        Random       & 144.84$\pm$12.45 & 68.94$\pm$2.33 \\ \midrule
        IS           & 139.83$\pm$9.87$\phantom{2}$  & 67.31$\pm$2.34 \\
        LPIPS        & 130.16$\pm$10.23 & 70.89$\pm$2.46 \\ \midrule
        ReLIC        & 122.25$\pm$7.04$\phantom{2}$  & 75.67$\pm$3.06 \\
        TANet        & 114.67$\pm$8.93$\phantom{2}$  & 74.23$\pm$2.93 \\
        SD-Chad      & 108.93$\pm$9.56$\phantom{2}$  & 77.65$\pm$1.86 \\
        VLAD         & 127.25$\pm$8.84$\phantom{2}$  & 76.28$\pm$2.24 \\ \midrule
        \NAME{} & \textbf{94.32$\pm$2.56}$\phantom{2}$ & \textbf{88.06$\pm$0.99} \\
    \bottomrule
    \end{tabular}}
    \label{tab:single_score_comparison}
\end{table}

\begin{figure}[t]
    \vspace{-0.5em}
    \centering
    \subfigure[Dynamic Schedule]{\includegraphics[width=0.45\linewidth]{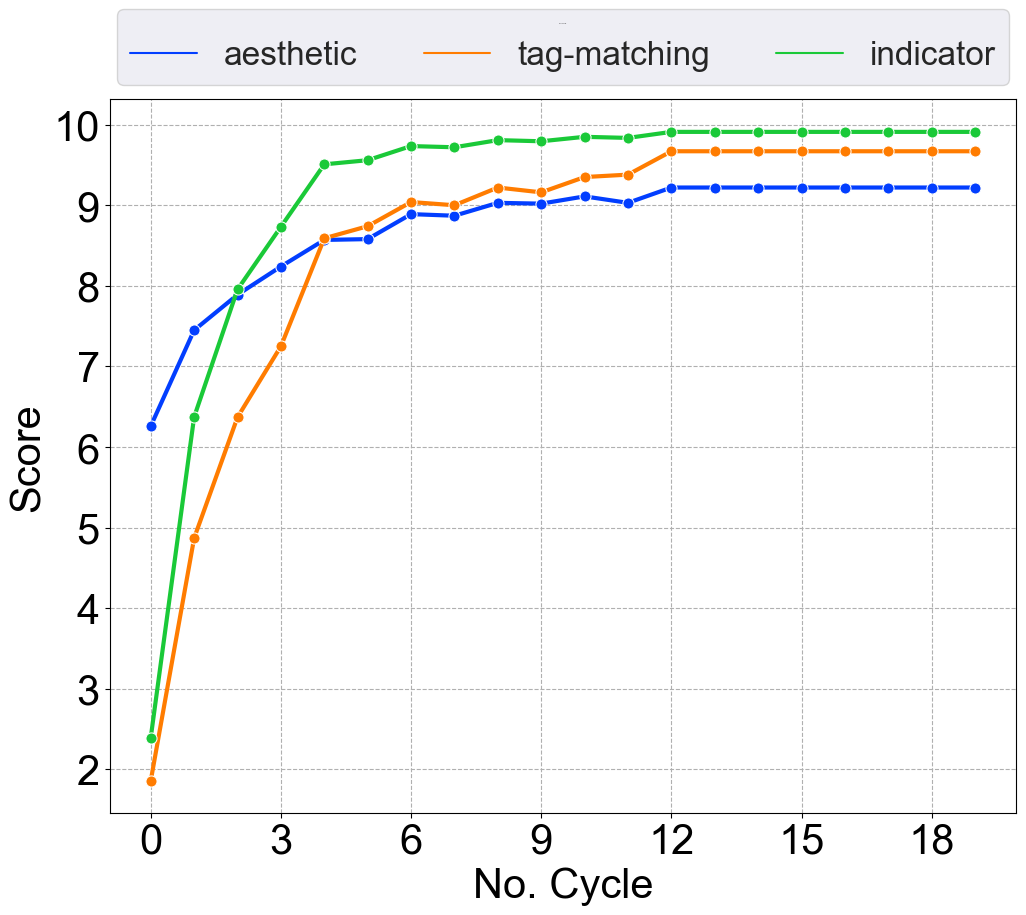}}
    \subfigure[Fixed Schedule]{\includegraphics[width=0.45\linewidth]{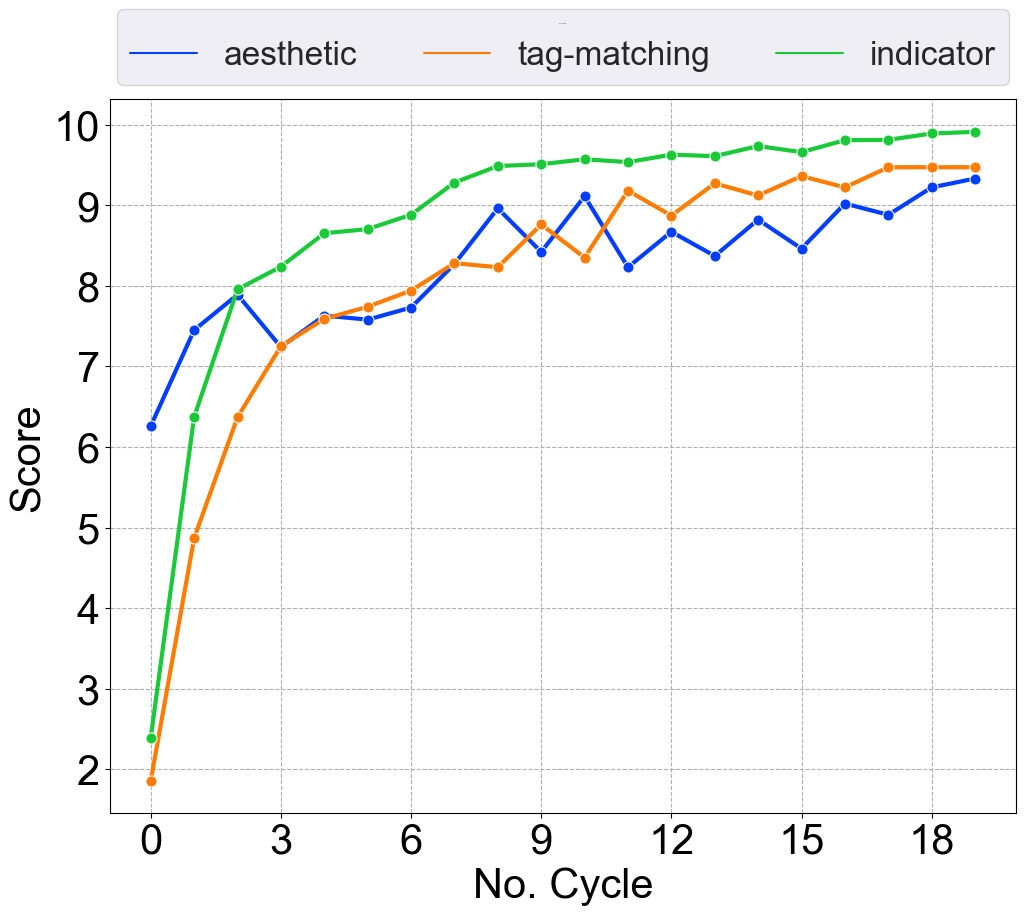}}
    \caption{\textbf{How the scores change on \NAME{} as cycle proceeds on emperor penguin(chick).} (a), (b) shows how aesthetic/concept-matching/comprehensive scores change across different cycles on \NAME{} with or without a dynamic schedule, respectively.
    }
    \label{fig:schedule_score_compare}
\end{figure}

\paragraph{Scores across cycles.} Fig.~\ref{fig:schedule_score_compare}(b) shows how the aesthetic/concept-matching/indicator score changes as the cycle proceeds. We can see that \NAME{} already achieves a higher value on all scores and cannot be further boosted at cycle 12, since more newly-added samples do not help boost embedding quality, making it stops updating and refining.
Furthermore, we also notice that aesthetic/concept-matching scores fluctuate at later cycles when the comprehensive score is enhanced continuously, while Fig.~\ref{fig:schedule_score_compare}(b) shows a more severe fluctuation without assistance from dynamic scheduling.
These phenomena exactly correspond to our proposed dual scoring system that balances the importance of these two factors, when ensuring the quality of embedding never declines.

\subsection{Ablation Studies}\label{exp:ablation}

We verify the effectiveness of all components in our proposed \NAME{} framework, including the scoring modules within the weighted scoring system, and the dynamic schedule designed according to the indicator $\phi$ in Eq.~\ref{eq:embedding_final_score}.
All the experiments in this section are conducted on ``emperor penguin chick''. The results are shown as follows.

\paragraph{Comparing with other single-score-based selection.}    First, we validate the effectiveness of our dual scoring system by comparing it with single-score-based selection in Table~\ref{tab:single_score_comparison}, which includes evaluation scores adopted from image-generation works and aesthetic/concept-matching scores. Specifically, generative-evaluation scores includes inception score~(IS)~\cite{salimans2016improved}, and learned perceptual image patch similarity~(LPIPS)~\cite{zhang2018unreasonable}, while aesthetic/concept-matching scores include ReLIC~\cite{zhao2020representation}, TANet~\cite{he2022rethinking}, SD-Chad and VLAD~\cite{sivic2003video}.
All the experiments are equipped with dynamic scheduling in Sec.~\ref{sec:dynamic_scheduling} for a fair comparison.

In Table~\ref{tab:single_score_comparison}, we can see the aesthetic/retrieval-based strategies consistently outperform the scores induced by generation-based evaluation metrics. We account this for that  evaluation-based metrics focus on a generic generation scheme rather than focusing on a given concept, making them unsuitable for this setting. We can also see that the dual scoring strategy in \NAME{} consistently outperforms the single-score-based ones~(14.61 on FID and 9.61\% on R-precision with 200 training samples).
From these comparisons, we claim that designing a scoring system specially designed for personalized text-to-image generation is important, and our weighted scoring system exactly meets this requirement with a weighted image selection strategy that adjusts the scoring preference according to the embeddings.



\begin{figure}[t]
\centering
\vspace{-1em}
\includegraphics[width=0.8\linewidth]{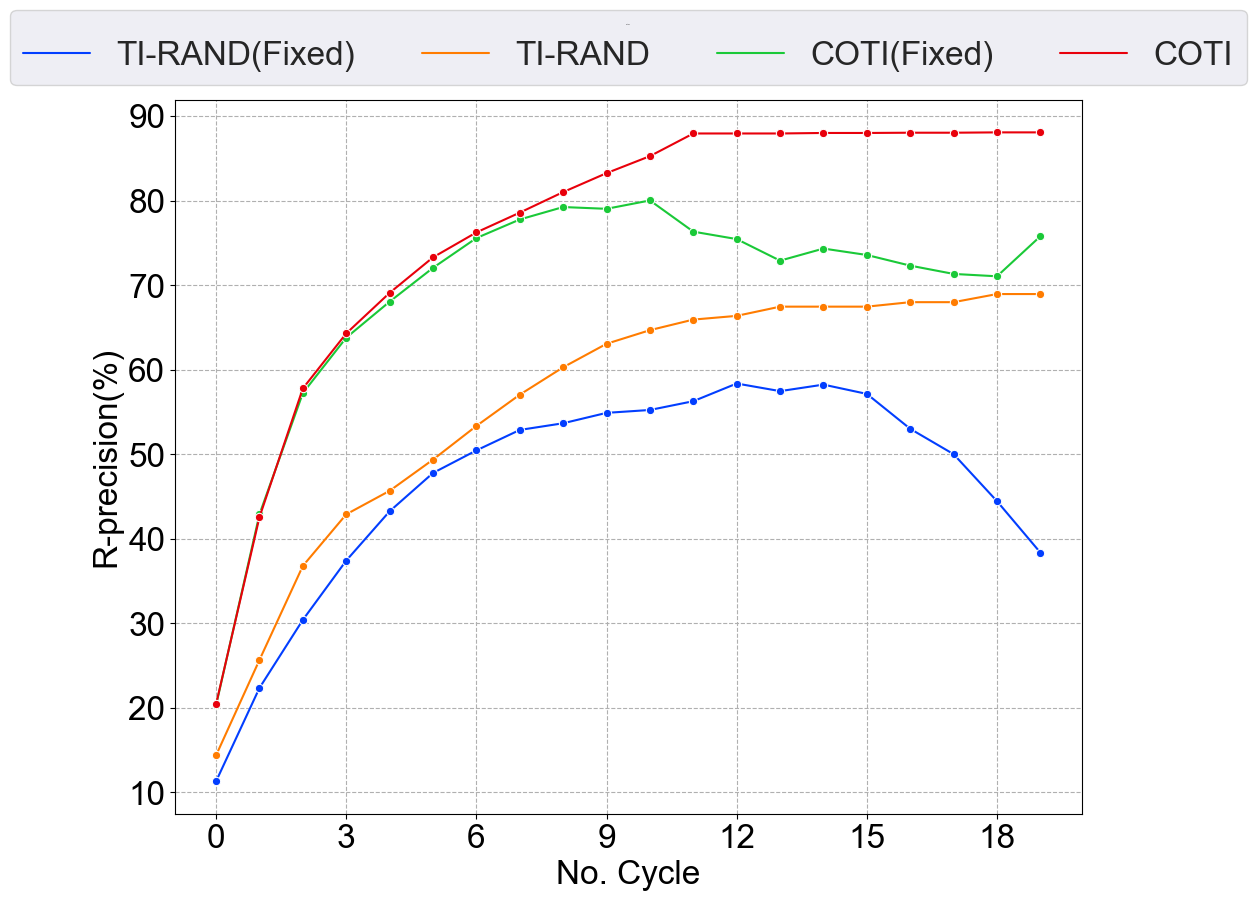}
\caption{\textbf{R-precision (\%) with different strategies as training pool expands.} The experiments are performed over "emperor penguin chick" and produced over TI(RAND) and \NAME{}. To further show the impact of the number of samples, we also compare their modified versions that do not use dynamic schedules.}
\label{fig:on_num_samples}
\end{figure}

\paragraph{The effect of dynamic scheduling.} 
To look at the influence of the dynamic schedule, we also compare the performance of TI (RAND) and \NAME{} with their schedule-fixed versions in Figure~\ref{fig:on_num_samples}. 
We can see that without a dynamic schedule, both TI (RAND) and \NAME{} faces performance loss or fluctuation with much more samples (i.e., 100 samples for \NAME{} and 120 samples for TI (RAND)). For TI (RAND), more samples do not necessarily give precise visual descriptions for a concept, which in turn leads to inferior performance compared to other sampling methods. For \NAME{}, samples at later cycles tend to have lower quality~(i.e. lower resolution, disturbing elements), which can include redundant or even harmful details. This problem can be alleviated via dynamic schedules. To conclude, the dynamic schedule coupled with the weighted scorer is also essential for training high-quality embeddings.


\section{Related work}


\paragraph{Personalized Text-to-image Synthesis.} The task of text-to-image generation refers to generate specific images based on text descriptions~\cite{bayoumi2021text,xia2022gan}, and has attained amazing performance with state-of-the-art diffusion models~\cite{ramesh2022hierarchical, saharia2022photorealistic, nichol2021glide, rombach2022high}.
Therefore, adapting large-scale text-to-image models to a specific concept while also preserving this amazing performance, i.e. \textit{personalization}~\cite{cohen2022my}, has become another recent research interest.
But this is often difficult, since re-training a model with an expanded dataset for each new concept is prohibitively expensive, while ﬁne-tuning the whole model~\cite{ding2022don,li2022overcoming} or transformation modules~\cite{zhou2022learning, gao2021clip, skantze2022collie} on few examples typically leads to some rate of forgetting~\cite{kumar2022fine,cohen2022my}.
More recently, \textit{textual inversion}~(TI)~\cite{gal2022image} tried to resolve these problems by representing the newly-given concept with pseudo word~\cite{rathvon2004early} and remapping it to another carefully trained embedding in the text-encoding space, guided by few images. 
In our framework, we follow the textual inversion approach, as it provides a fast approach to represent and generate new objects with relatively low costs. 

\paragraph{Active Learning.} Active Learning is a machine learning paradigm that actively selects the data best for training models from external data sources~\cite{ren2021survey}.
The most crucial part of active learning is exactly the data acquisition, i.e. the strategy to select the optimal data batch.
Current studies can be roughly categorized as follows: (a) Score-based methods that prefer the samples with the highest information scores~\cite{mai2022effectiveness,wang2022boosting}; (b) Representation-based methods searching for the samples that are the most representative of the underlying data distribution~\cite{DBLP:conf/iclr/SenerS18,DBLP:conf/iclr/AshZK0A20}.

In this work, we focus on scored-based strategies in two-fold: \textit{aesthetics} and \textit{concept-matching}.
Image Aesthetics Assessment~(IAA) aims at evaluating image aesthetics computationally and automatically~\cite{deng2017image}, while automatically assessing image aesthetics is useful for many applications~\cite{luo2008photo,kong2016photo,fang2020perceptual}. To take a step further, personalized image aesthetics assessment~(PIAA)~\cite{ren2017personalized,zhu2021learning,yang2022personalized} was proposed to capture unique aesthetic preferences, consistent with our goal to adjust our paradigms accordingly different concepts.
At the same time, "Concept-matching" is adopted from the field of image retrieval, in which we search for relevant images in an image gallery by analyzing the visual content (e.g., objects, colors, textures, shapes \textit{etc}.), given a query image~\cite{smeulders2000content, lew2006content}.
To design a paradigm that automatically meets the harsh requirements for embedding training, we take both two factors into consideration within our scorer.

\section{Conclusion}

We propose \NAME{}, an enhanced, data-efficient, and practically useful version of text inversion for AIGC.
This method is encapsulated in an active-learned paradigm with carefully designed acquisitional scoring mechanisms.
\NAME{} significantly outperforms the prior TI-related approaches in many (if not most) aspects including generation quality, concept matches, technological robustness, data efficiency, etc.
In the future, we hope to ship \NAME{} to the open-source community so as to absorb more long-tailed concept embeddings that are originally uncovered, via the assistance of collective efforts.


{\small
\bibliographystyle{ieee_fullname}
\bibliography{egbib}
}

\end{document}